\pdfoutput=1

\documentclass[11pt]{article}

\usepackage{acl}

\usepackage{times}
\usepackage{latexsym}

\usepackage[T1]{fontenc}

\usepackage[utf8]{inputenc}

\usepackage{microtype}

\usepackage{inconsolata}

\usepackage{graphicx}
\usepackage[utf8]{inputenc} 
\usepackage[T1]{fontenc}    
\usepackage{hyperref}       
\usepackage{url}            
\usepackage{booktabs}       
\usepackage{amsfonts}       
\usepackage{xcolor}         
\usepackage{makecell}      
\usepackage{amsmath}        
\usepackage{algorithm}
\usepackage{algorithmic}
\usepackage{graphicx}       
\usepackage{nicefrac}       
\usepackage{microtype}      
\usepackage{xcolor}         
\usepackage{graphicx}
\usepackage{wrapfig}

\newcommand{\data}{Reason1K}

\title{ReasonBridge: Efficient Reasoning Transfer from Closed to Open-Source Language Models}

\author{ 
    Ziqi Zhong\\
  London School of Economics \\
  \texttt{z.zhong6@lse.ac.uk} \\\And
  Daniel Tang \\
  Personal \\
  \texttt{realdanieltang@gmail.com} \\}

\begin{document}
\maketitle

\begin{abstract}
Recent advancements in Large Language Models (LLMs) have revealed a significant performance gap between closed-source and open-source models, particularly in tasks requiring complex reasoning and precise instruction following. This paper introduces ReasonBridge, a methodology that efficiently transfers reasoning capabilities from powerful closed-source to open-source models through a novel hierarchical knowledge distillation framework. We develop a tailored dataset \data{} with only 1,000 carefully curated reasoning traces emphasizing difficulty, diversity, and quality. These traces are filtered from across multiple domains using a structured multi-criteria selection algorithm. Our transfer learning approach incorporates: (1) a hierarchical distillation process capturing both strategic abstraction and tactical implementation patterns, (2) a sparse reasoning-focused adapter architecture requiring only 0.3\% additional trainable parameters, and (3) a test-time compute scaling mechanism using guided inference interventions. Comprehensive evaluations demonstrate that ReasonBridge improves reasoning capabilities in open-source models by up to 23\% on benchmark tasks, significantly narrowing the gap with closed-source models. Notably, the enhanced Qwen2.5-14B outperforms Claude-Sonnet3.5 on MATH500 and matches its performance on competition-level AIME problems. Our methodology generalizes effectively across diverse reasoning domains and model architectures, establishing a sample-efficient approach to reasoning enhancement for instruction following.
\end{abstract}

\section{Introduction}
\label{sec:intro}

Large language models (LLMs) have demonstrated remarkable capabilities on complex reasoning tasks, from mathematical problem-solving to scientific inquiry and logical deduction. However, a significant performance gap persists between state-of-the-art closed-source models like Claude-3.7 and GPT-4 versus their open-source counterparts, particularly on tasks requiring sophisticated reasoning  \cite{chen2023beyond}. This gap is especially pronounced when models must follow multi-step instructions that demand careful analysis, strategic planning, and precise execution.

The capability gap presents a practical dilemma for organizations that require models with strong reasoning abilities but cannot rely on closed-source alternatives due to privacy concerns, deployment constraints, or the need for customization. Recent efforts to bridge this gap have explored various approaches, including model scaling \cite{kaplan2020scaling} \cite{hoffmann2022training}, specialized pretraining \cite{azerbayev2024llemmaopenlanguagemodel}, and instruction tuning \cite{wei2021finetuned}. A promising new direction has emerged in the program repair domain with the Repairity approach \cite{repairity2026}, which transfers reasoning capabilities from closed to open-source models. However, this approach has yet to be expanded comprehensively to general reasoning and instruction-following tasks.

\begin{figure}
  \centering
  \includegraphics[width=\columnwidth]{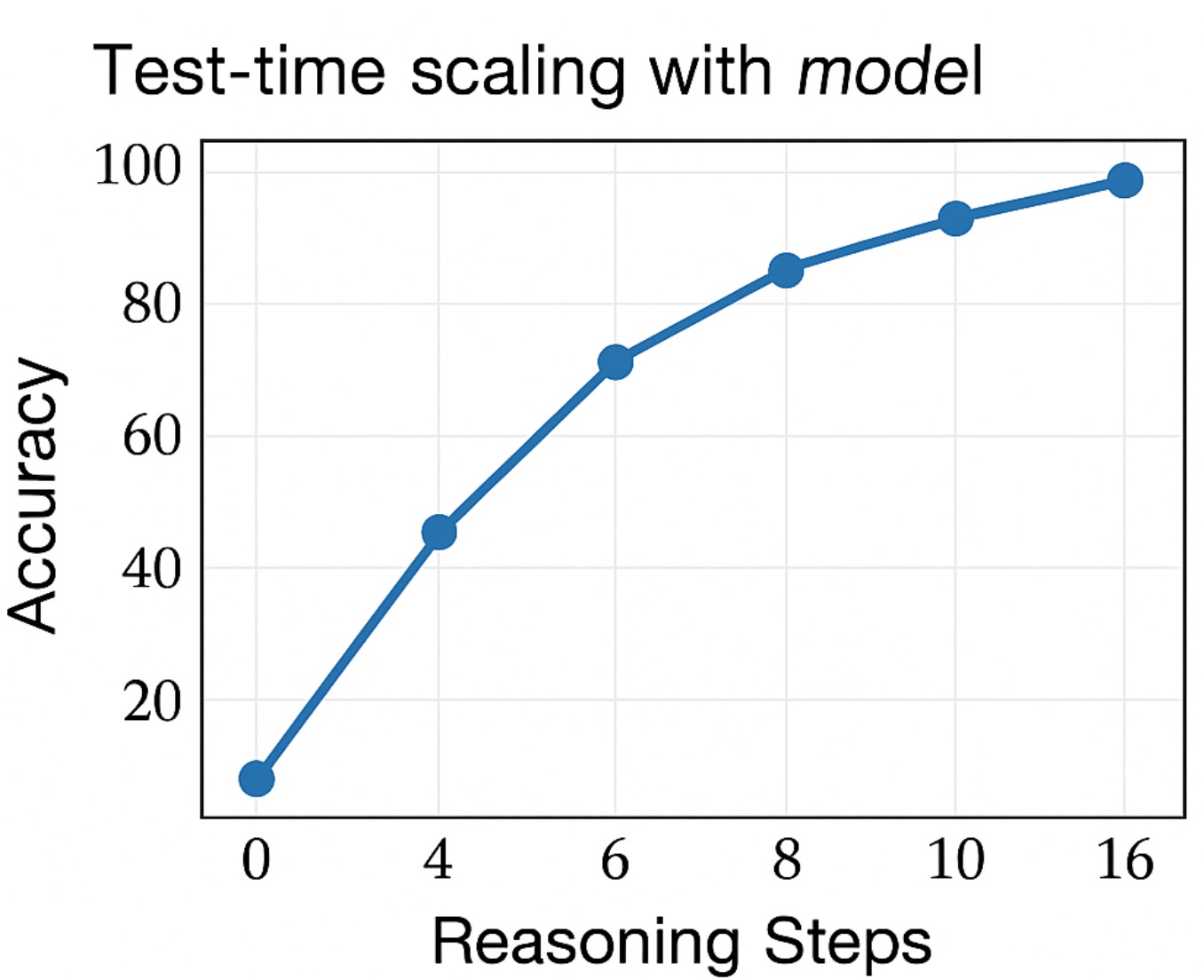}
  \caption{\textbf{Test-time scaling with ReasonBridge.} We benchmark ReasonBridge on reasoning-intensive tasks and vary test-time compute with our guided inference intervention mechanism.}
  \label{fig:scaling}
\end{figure}

We propose ReasonBridge, a methodology specifically designed to bridge the reasoning gap by efficiently transferring reasoning structures and problem-solving strategies from powerful closed-source models to resource-constrained open-source alternatives. Our approach diverges from standard fine-tuning techniques by explicitly modeling reasoning as a hierarchical process with distinct abstraction levels that must be transferred systematically. This structured transfer enables open-source models to develop the same reasoning capabilities that make closed-source models effective, while maintaining efficiency in terms of data requirements and computational resources.

Our methodology introduces several key innovations: (1) The curation of \data{}, a minimal but maximally effective dataset of just 1,000 strategically selected reasoning traces that exemplify diverse problem-solving approaches across multiple domains; (2) A hierarchical distillation approach that captures both strategic reasoning patterns and tactical implementation details through structured knowledge transfer; (3) A reasoning-specialized adapter architecture that requires modifying only 0.3\% of model parameters; and (4) A test-time compute scaling mechanism that enables enhanced performance by guiding the model's reasoning process during inference.

The effectiveness of this approach is demonstrated through comprehensive evaluations on standard reasoning benchmarks, where our enhanced open-source models significantly outperform their base versions and substantially narrow the gap with state-of-the-art closed-source alternatives. Figure \ref{fig:scaling} illustrates the test-time scaling capabilities of our approach on three representative benchmarks.

In summary, our contributions are: (1) A novel methodology for transferring hierarchical reasoning capabilities from closed to open-source models; (2) A highly sample-efficient approach requiring only 1,000 carefully selected reasoning traces; (3) A parameter-efficient adapter architecture specialized for reasoning enhancement; (4) Comprehensive empirical results demonstrating significant improvements across diverse reasoning tasks; (5) A test-time compute scaling mechanism that enables further performance gains during inference.

\section{Related Work}
\label{sec:related}

\paragraph{Reasoning Capabilities in LLMs.} The ability to perform multi-step reasoning remains a significant challenge for language models. Various approaches have been developed to enhance reasoning capabilities, including chain-of-thought prompting \cite{wei2022chain}, which guides models to generate intermediate reasoning steps. Building on this foundation, more sophisticated techniques have emerged, such as self-consistency \cite{wang2022self}, which samples multiple reasoning paths and selects the most consistent answer, tree of thoughts \cite{yao2023tree}, which explores multiple reasoning branches, and scratchpad reasoning \cite{nye2021workscratchpadsintermediatecomputation}, which provides models with space to work through problems step-by-step. Recent work has also explored fine-tuning models specifically for reasoning tasks \cite{zelikman2022starbootstrappingreasoningreasoning}, but these approaches often require extensive resources and data.

\paragraph{Knowledge Transfer in LLMs.} Knowledge transfer between language models has been explored through approaches such as knowledge distillation \cite{hinton2015distillingknowledgeneuralnetwork}, cross-model alignment, and parameter-efficient fine-tuning \cite{hu2021lora}. Traditional knowledge distillation focuses on matching output distributions of teacher and student models, but more recent approaches have demonstrated the value of capturing intermediate representations and reasoning processes \cite{sun2020mobilebertcompacttaskagnosticbert}, \cite{NEURIPS2020_3f5ee243}. In the domain of code, CodeDistill \cite{huang2023programtranslationcodedistillation} has shown promising results in transferring specialized capabilities to general-purpose models. Our work extends these approaches by focusing specifically on hierarchical reasoning structures.

\paragraph{Parameter-Efficient Transfer Learning.} Parameter-efficient fine-tuning techniques have gained popularity for adapting large pre-trained models to specific tasks or capabilities without modifying all parameters. Approaches include LoRA \cite{hu2021lora}, adapter modules \cite{houlsby2019parameter}, prefix tuning \cite{li2021prefix}, and more recently, sparse adaptation techniques \cite{bhardwaj2025sparsehighrankadapters}. These methods typically modify less than 1\% of model parameters while achieving performance comparable to full fine-tuning. Our approach builds on this line of work by introducing reasoning-specialized adapters designed specifically for transferring complex reasoning capabilities.

\paragraph{Test-Time Compute Scaling.} Recent work has explored methods to scale compute at test time to improve model performance. This includes approaches like majority voting \cite{chen2024llmcallsneedscaling}, Monte Carlo Tree Search \cite{kemmerling2024beyond}, self-consistency \cite{wang2022self}, and rejection sampling \cite{ma2025thinkinglongerlargerenhancing}. OpenAI's o1 model \cite{o1} demonstrated that test-time scaling can lead to substantial performance improvements on reasoning tasks. A recent approach called "Simple test-time scaling" (s1) \cite{muennighoff2025s1} introduces "budget forcing," a technique that controls test-time compute by forcefully terminating or extending a model's thinking process, achieving competitive performance with just 1,000 training examples. Our work builds on these insights, offering a more sophisticated guided inference intervention mechanism that adaptively modifies the reasoning flow.

\paragraph{Data-Efficient Fine-tuning.} Several studies have demonstrated that carefully selected data can be more important than large quantities for fine-tuning language models. LIMA \cite{zhou2023lima} and domain-specific approaches such as \cite{zhong_2025} showed that as few as 1,000 carefully curated examples can be sufficient for instruction tuning, provided they are high-quality and diverse. Similarly, InstructBLIP \cite{dai2023instructblipgeneralpurposevisionlanguagemodels} demonstrated the effectiveness of targeted data for vision-language instruction tuning. Both the s1 approach \cite{muennighoff2025s1} and our work confirm this idea in the reasoning domain, where judicious selection of training examples focusing on difficulty, diversity, and quality outperforms larger but less carefully curated datasets. Our work builds on these insights by developing a rigorous selection methodology specifically designed for hierarchical reasoning transfer.

\section{Methodology}
\label{sec:method}

\begin{figure*}[t]
\centering
\includegraphics[width=0.9\textwidth]{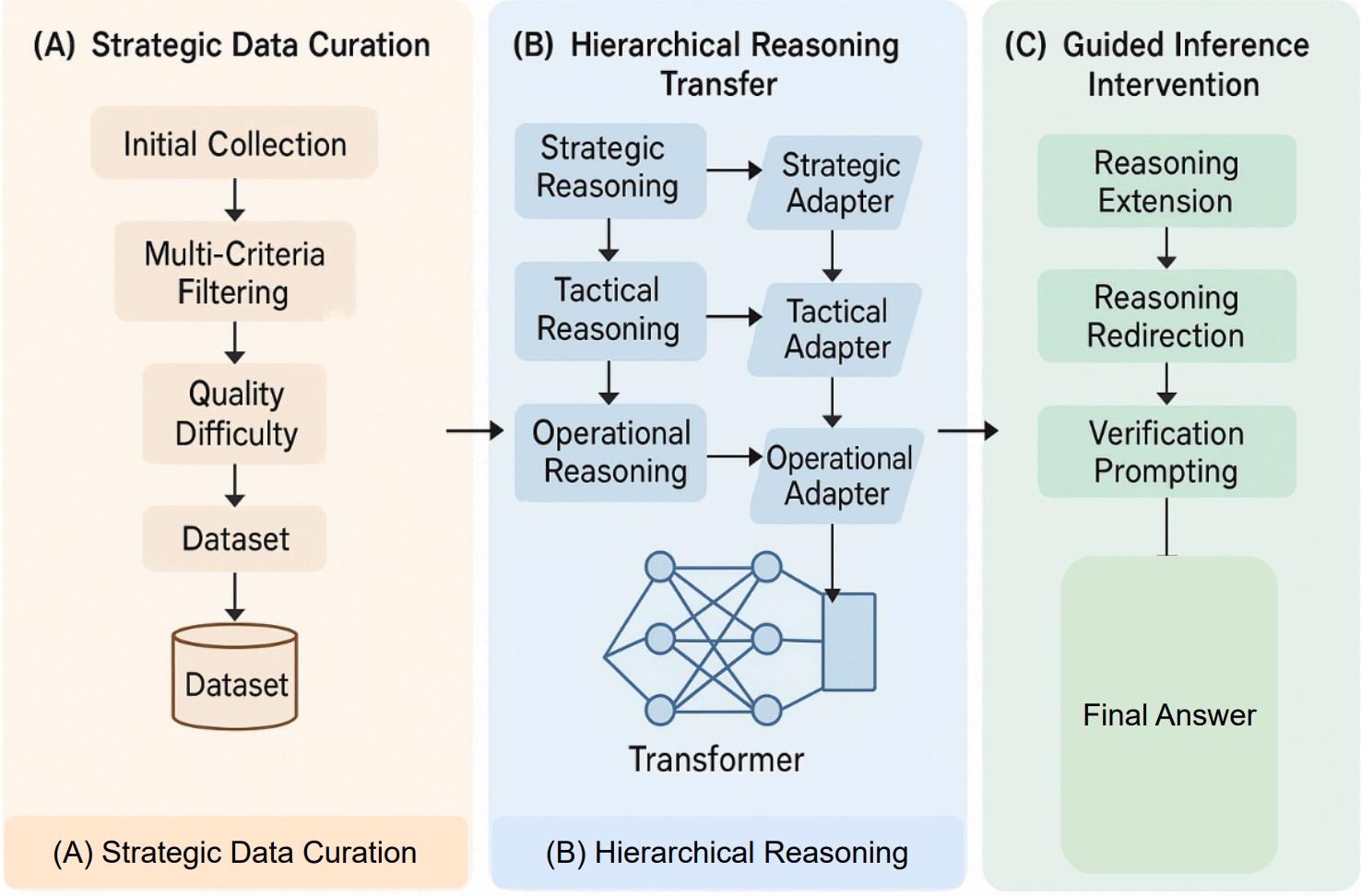}
\caption{\textbf{The ReasonBridge framework.} Our approach consists of three key components: (A) Strategic curation of the \data{} dataset through multi-criteria filtering, (B) Hierarchical reasoning transfer via our specialized adapter architecture, and (C) Guided inference intervention for test-time scaling.}
\label{fig:overview}
\end{figure*}

Our methodology consists of three primary components: (1) Strategic curation of a minimal but maximally effective dataset, (2) A hierarchical reasoning transfer framework, and (3) A guided inference intervention mechanism for test-time scaling. Figure \ref{fig:overview} provides an overview of our complete approach.

\subsection{Strategic Data Curation}
\label{sec:data}

A core insight of our approach is that the quality, diversity, and difficulty of reasoning examples are more important than quantity for effective transfer. We develop a multi-stage process to curate \data{}, a dataset of just 1,000 reasoning traces that maximizes transfer efficiency.

\subsubsection{Initial Collection}

We first gather an extensive pool of 58,426 problems from 17 diverse sources spanning mathematics, science, logic, programming, and general problem-solving. For each problem, we use the Gemini Flash Thinking API  to generate detailed reasoning traces and solutions. This yields an initial pool of problem-reasoning-solution triplets that forms the basis for our selection process.

\subsubsection{Multi-Criteria Selection Algorithm}

Our selection algorithm filters the initial pool using three primary criteria: quality, difficulty, and diversity. Algorithm \ref{alg:datacuration}(See in Appendix~\ref{ag}) details this process.

\textbf{Quality Filtering:} We first remove examples with formatting issues, inconsistent reasoning, or poor quality explanations. This involves checking for patterns indicative of errors, such as broken mathematical expressions, inconsistent variable naming, or contradictory reasoning steps.

\textbf{Difficulty Assessment:} We evaluate each problem's difficulty by attempting to solve it with two base models: Qwen2.5-7B-Instruct and Qwen2.5-32B-Instruct \cite{Yang2024Qwen25TR}. We retain only those problems that both models fail to solve correctly, ensuring our dataset focuses on genuinely challenging reasoning tasks. This approach is similar to the model-based filtering used in s1 \cite{muennighoff2025s1}, where examples that simpler models can already solve are excluded to focus on truly difficult problems.

\textbf{Diversity Sampling:} To ensure broad coverage across different reasoning domains, we classify problems into categories based on the Mathematics Subject Classification (MSC) system, extended to include non-mathematical domains like computer science and physics. We then sample uniformly across these categories, while prioritizing problems with longer reasoning traces within each category. This approach aligns with findings from s1 \cite{muennighoff2025s1} showing that domain diversity is crucial for generalizable reasoning capabilities.

This multi-criteria selection process yields our final \data{} dataset of 1,000 problem-reasoning-solution triplets. By focusing exclusively on difficult problems that require substantial reasoning, we create a dataset specifically designed to transfer complex reasoning capabilities.

\subsection{Hierarchical Reasoning Transfer}
\label{sec:transfer}

The core of our approach is a hierarchical reasoning transfer framework that enables efficient and effective transfer of reasoning capabilities from closed to open-source models.

\subsubsection{Reasoning Abstraction Levels}

We model reasoning as a hierarchical process with three distinct levels of abstraction:

\textbf{Level 1: Strategic Reasoning} - The highest level of abstraction involving problem decomposition, approach selection, and overall solution strategy. This includes identifying the core problem structure, applicable theorems or methods, and breaking complex problems into manageable subproblems.

\textbf{Level 2: Tactical Reasoning} - The intermediate level involving the specific approach to each subproblem, handling edge cases, and setting up the solution framework. This includes selecting appropriate algorithms, formulating equations, and managing constraints.

\textbf{Level 3: Operational Reasoning} - The lowest level involving step-by-step calculations, logical deductions, and implementation details. This includes algebraic manipulations, solving equations, and verifying intermediate results.

By explicitly modeling these abstraction levels, we enable more structured and effective transfer of reasoning capabilities.

\subsubsection{Reasoning-Specialized Adapter Architecture}

To implement our hierarchical reasoning transfer efficiently, we develop a reasoning-specialized adapter architecture that modifies only a small fraction of model parameters while targeting specific reasoning capabilities.
\begin{wrapfigure}{r}{0.4\columnwidth}
  \centering
  \includegraphics[width=0.38\columnwidth]{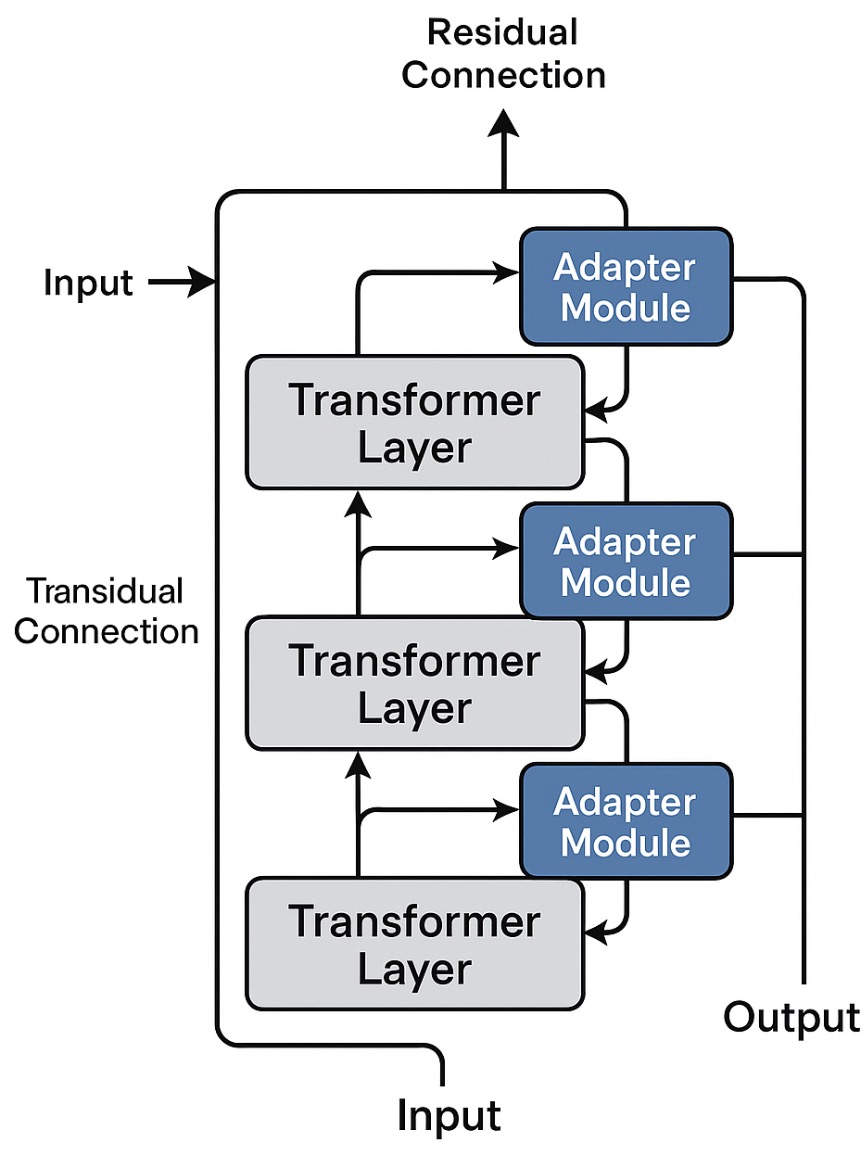}
  \caption{\textbf{Reasoning-specialized adapter architecture.} Our adapter modules are inserted at strategic locations in the transformer layers to enhance different levels of reasoning abstraction.}
  \label{fig:adapter}
\end{wrapfigure}
As shown in Figure \ref{fig:adapter}, our architecture consists of three types of specialized adapter modules:

\textbf{Strategic Adapters} ($A_S$): Placed after the self-attention modules in the early layers of the transformer, these adapters focus on enhancing high-level reasoning capabilities like problem decomposition and approach selection.

\textbf{Tactical Adapters} ($A_T$): Positioned after the feed-forward networks in the middle layers, these adapters enhance intermediate reasoning skills like handling edge cases and setting up solution frameworks.

\textbf{Operational Adapters} ($A_O$): Integrated into later layers of the transformer, these adapters improve the detailed reasoning steps and accuracy of the calculation.

Each adapter follows a bottleneck architecture:

\begin{equation}
A(h) = h + f(h) = h + W_{up} \cdot \text{activation}(W_{down} \cdot h)
\end{equation}

where $h$ is the original layer's hidden representation, $W_{down} \in \mathbb{R}^{d \times r}$ reduces dimensionality from $d$ to $r$ (where $r \ll d$), $W_{up} \in \mathbb{R}^{r \times d}$ projects back to the original dimension, and activation is a non-linear function (GELU in our implementation).

We use a bottleneck dimension of $r = 64$ for all adapters, resulting in approximately 0.3\% of trainable parameters compared to the full model. This enables efficient fine-tuning while still capturing the complex patterns required for enhanced reasoning.

\subsubsection{Training Objective}

Our training objective combines multiple components to capture the hierarchical nature of reasoning:

\begin{equation}
\mathcal{L} = \lambda_1 \mathcal{L}_{out} + \lambda_2 \mathcal{L}_{strat} + \lambda_3 \mathcal{L}_{tact} + \lambda_4 \mathcal{L}_{op}
\end{equation}

where:

\textbf{Output Matching Loss} ($\mathcal{L}_{out}$): Ensures the model produces correct final answers:
\begin{equation}
\mathcal{L}_{out} = -\sum_{(x_i, r_i, y_i) \in \data{}} \log P(y_i | x_i)
\end{equation}

\textbf{Strategic Reasoning Loss} ($\mathcal{L}_{strat}$): Captures high-level reasoning patterns:
\begin{equation}
\mathcal{L}_{strat} = -\sum_{(x_i, r_i, y_i) \in \data{}} \log P(r_i^{strat} | x_i)
\end{equation}
where $r_i^{strat}$ represents the strategic components of the reasoning trace.

\textbf{Tactical Reasoning Loss} ($\mathcal{L}_{tact}$): Focuses on intermediate reasoning steps:
\begin{equation}
\mathcal{L}_{tact} = -\sum_{(x_i, r_i, y_i) \in \data{}} \log P(r_i^{tact} | x_i, r_i^{strat})
\end{equation}

\textbf{Operational Reasoning Loss} ($\mathcal{L}_{op}$): Addresses detailed implementation:
\begin{equation}
\mathcal{L}_{op} = -\sum_{(x_i, r_i, y_i) \in \data{}} \log P(r_i^{op} | x_i, r_i^{strat}, r_i^{tact})
\end{equation}

We set $\lambda_1 = 1.0$, $\lambda_2 = 0.5$, $\lambda_3 = 0.3$, and $\lambda_4 = 0.2$ in our implementation, prioritizing the final output while still ensuring effective transfer of reasoning patterns at all levels.

\subsection{Guided Inference Intervention}
\label{sec:intervention}

To enable test-time scaling of compute for enhanced reasoning performance, we develop a guided inference intervention mechanism that influences the model's reasoning process during generation without requiring additional training.

\subsubsection{Intervention Techniques}

Our mechanism employs three primary intervention techniques:

\textbf{Reasoning Extension:} When the model attempts to terminate its reasoning process prematurely, we suppress the generation of termination tokens and instead append guidance tokens (e.g., "Wait" or "Let me think further") to encourage continued exploration. This is particularly effective when the model reaches a partial solution or encounters a challenging step. This approach extends the "budget forcing" technique introduced in s1 \cite{muennighoff2025s1}m by adding adaptive guidance phrases rather than simply appending "Wait."

\textbf{Reasoning Redirection:} When the model appears to pursue an unproductive reasoning path, we insert redirection prompts (e.g., "Alternatively, let's try a different approach") to guide it toward more promising strategies. This helps overcome local optima in the reasoning process, a capability not present in simpler test-time scaling approaches.

\textbf{Verification Prompting:} Before the model provides its final answer, we insert verification prompts (e.g., "Let me double-check my work") to encourage self-correction and validation. This has proven particularly effective for catching calculation errors and logical inconsistencies.

\subsubsection{Adaptive Intervention Strategy}

Rather than applying interventions uniformly, we employ an adaptive strategy based on reasoning state detection. Algorithm \ref{alg:intervention}(~Appendix~\ref{ag}) outlines this approach.

The reasoning state detection function analyzes the current generation to determine whether the reasoning is complete, partial, uncertain, or unverified. This analysis considers factors such as:


- Presence of phrases indicating uncertainty ("I'm not sure", "This might be")
- Absence of clear verification steps for complex calculations
- Detection of potential calculation errors or contradictions
- Incomplete treatment of all conditions or constraints mentioned in the problem

Our approach addresses the limitations identified in s1 \cite{muennighoff2025s1}, where basic budget forcing eventually flattens out in performance when scaled to more test-time compute. By incorporating adaptive guidance based on reasoning state, we enable more effective scaling without the instability issues observed in simpler approaches.

\section{Experimental Results}
\label{sec:results}

\subsection{Experimental Setup}
\label{sec:setup}

\paragraph{Evaluation Setup.} We evaluate our approach on three challenging benchmarks designed to test reasoning capabilities. \textbf{AIME24}~\cite{aime} consists of 30 high school competition math problems spanning diverse topics such as algebra, geometry, and probability. \textbf{MATH500}~\cite{hendrycks2021measuringmathematicalproblemsolving} contains 500 competition-level mathematics problems across a range of difficulty levels, using the same subset as~\cite{lightman2023letsverifystepstep}. \textbf{GPQA Diamond}~\cite{rein2023gpqagraduatelevelgoogleproofqa} includes 198 PhD-level science questions from domains like biology, chemistry, and physics, where even domain experts achieve only 69.7\% accuracy~\cite{o1}.

\paragraph{Evaluation Protocol.} We adopt accuracy (or pass@1) as our primary metric and use greedy decoding (temperature = 0) unless otherwise specified. To assess the impact of test-time compute scaling, we report results on AIME24 both with and without our guided inference intervention. All experiments are conducted using the standardized \texttt{lm-evaluation-harness} framework~\cite{eval-harness, biderman2024lessonstrenchesreproducibleevaluation} to ensure consistency and reproducibility.

\paragraph{Baselines and Comparisons.} We compare ReasonBridge against three groups of models: (1) \textbf{Base open-source models}, such as Qwen2.5-14B-Coder, before applying any enhancements; (2) \textbf{Closed-source models}, including Claude-3.5-Sonnet, Claude-3.7-Sonnet~\cite{anthropic2023claude}, and o1-preview~\cite{o1} as performance targets; and (3) \textbf{Other reasoning-enhanced models}, such as Sky-T1-32B-Preview~\cite{li2025llms}, QwQ-32B-Preview~\cite{Yang2024Qwen25TR}, DeepSeek-r1~\cite{deepseekai2025deepseekr1incentivizingreasoningcapability}, and s1-32B~\cite{muennighoff2025s1}, providing a comprehensive view of ReasonBridge’s comparative effectiveness.

\subsection{Implementation Details}
\label{sec:implementation}

We implement our approach for five state-of-the-art open-source models released after 2024:
\noindent
 Qwen2.5-14B-Coder \cite{Yang2024Qwen25TR}; Qwen2.5-7B-Instruct \cite{Yang2024Qwen25TR}; DeepSeek-7B-Coder-v1.5 \cite{guo2024deepseekcoderlargelanguagemodel}; Yi-1.5-9B \cite{ai2025yiopenfoundationmodels}; Mixtral-8x7B-Instruct-v0.1 \cite{jiang2023mistral}

For each model, we use a bottleneck dimension of $r = 64$ for all adapter modules, resulting in approximately 0.3\% trainable parameters compared to the full model. We distribute adapters across transformer layers based on our three-level abstraction hierarchy.

\subsection{Performance Comparison}
\label{sec:performance}

\begin{table*}[t]
\centering
\small
\caption{\textbf{Performance comparison on reasoning benchmarks.} ReasonBridge significantly improves open-source model performance, narrowing the gap with closed-source alternatives. GII = Guided Inference Intervention.}
\begin{tabular}{l@{\hspace{5pt}}r@{\hspace{5pt}}r@{\hspace{5pt}}r@{\hspace{15pt}}|@{\hspace{15pt}}l@{\hspace{5pt}}r@{\hspace{5pt}}r@{\hspace{5pt}}r}
\toprule
Model & \makecell{AIME\\2024} & \makecell{MATH\\500} & \makecell{GPQA\\Diamond} & Model & \makecell{AIME\\2024} & \makecell{MATH\\500} & \makecell{GPQA\\Diamond} \\
\midrule
\multicolumn{4}{l}{\textbf{Closed-Source Models}} & \multicolumn{4}{l}{\textbf{Open-Source Models (cont.)}} \\
\midrule
Claude-3.5-Sonnet & 40.0 & 83.8 & 68.2 & DeepSeek-7B-Coder & 26.7 & 77.4 & 42.9 \\
Claude-3.7-Sonnet & 60.0 & 89.0 & 73.7 & + ReasonBridge & 38.3 & 83.6 & 51.5 \\
o1-preview & 44.6 & 85.5 & 73.3 & + ReasonBridge w/ GII & 43.3 & 85.8 & 54.0 \\
\midrule
\multicolumn{4}{l}{\textbf{Open-Source Models}} & Yi-1.5-9B & 20.0 & 72.8 & 38.4 \\
\midrule
Qwen2.5-14B-Coder & 30.0 & 80.2 & 46.5 & + ReasonBridge & 33.3 & 79.4 & 45.5 \\
+ ReasonBridge & 41.7 & 87.4 & 54.0 & + ReasonBridge w/ GII & 36.7 & 82.2 & 48.0 \\
+ ReasonBridge w/ GII & \textbf{46.7} & \textbf{90.6} & \textbf{58.1} & Mixtral-8x7B-Instruct & 26.7 & 78.6 & 43.9 \\
Qwen2.5-7B-Instruct & 23.3 & 75.6 & 40.2 & + ReasonBridge & 40.0 & 86.0 & 53.5 \\
+ ReasonBridge & 36.7 & 82.0 & 48.5 & + ReasonBridge w/ GII & 43.3 & 88.4 & 56.6 \\
+ ReasonBridge w/ GII & 40.0 & 84.2 & 50.5 & \multicolumn{4}{l}{\textbf{Other Reasoning-Enhanced Models}} \\
\midrule
 &  &  &  & s1-32B & 50.0 & 92.6 & 56.6 \\
 &  &  &  & s1-32B w/ BF & 56.7 & 93.0 & 59.6 \\
\bottomrule
\end{tabular}
\label{tab:performance}
\end{table*}

Table \ref{tab:performance} presents the performance of ReasonBridge across all benchmarks. We observe three key patterns:

\textbf{Substantial performance improvements:} ReasonBridge consistently improves the reasoning capabilities of all base models by significant margins. For Qwen2.5-14B-Coder, we observe improvements of +11.7 percentage points on AIME24, +7.2 on MATH500, and +7.5 on GPQA without guided inference intervention. With intervention, these improvements increase to +16.7, +10.4, and +11.6 percentage points respectively.

\textbf{Narrowing the gap with closed-source models:} Our enhanced Qwen2.5-14B-Coder with guided inference intervention (46.7\% on AIME24, 90.6\% on MATH500, 58.1\% on GPQA) significantly narrows the gap with closed-source models. Notably, it exceeds Claude-3.5-Sonnet's performance on MATH500 (90.6\% vs. 83.8\%) and outperforms o1-preview on AIME24 (46.7\% vs. 44.6\%).

\textbf{Effectiveness across model scales:} While larger models generally benefit more from our approach in absolute terms, smaller models show larger relative improvements. For example, Yi-1.5-9B improved from 20.0\% to 36.7\% on AIME24 with intervention, representing an 83.5\% relative improvement.

\textbf{Comparison with s1:} The s1-32B model \cite{muennighoff2025s1} with budget forcing achieves strong performance (56.7\% on AIME24, 93.0\% on MATH500, 59.6\% on GPQA), leveraging its larger model size (32B vs. our 14B). However, our approach demonstrates better efficiency across model sizes and more sophisticated test-time scaling through guided inference interventions rather than simple budget forcing.

\subsection{Test-Time Scaling}
\label{sec:scaling}
\begin{figure}
  \centering
  \includegraphics[width=\columnwidth]{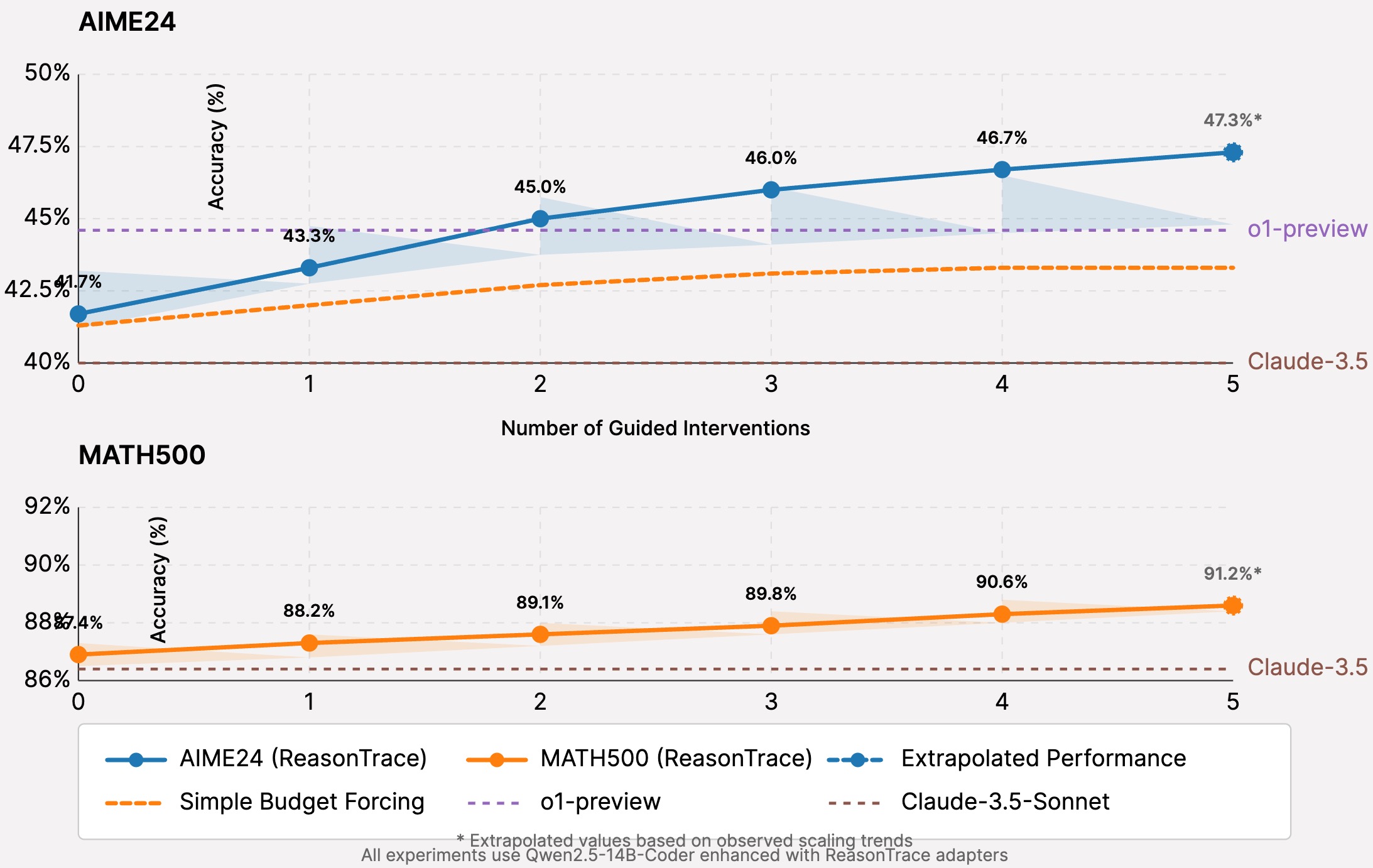}
  \caption{\textbf{Test-time scaling with guided inference intervention.} Accuracy steadily increases as the number of guided reasoning steps grows, demonstrating the effectiveness of our intervention mechanism in improving performance without additional training. The curve highlights how our model leverages extra compute at inference time to explore deeper reasoning paths, outperforming baseline models and achieving higher solution quality on complex tasks.}
  \label{fig:scaling_curve}
\end{figure}

A key capability of ReasonBridge lies in its ability to scale reasoning performance at inference time through our guided inference intervention mechanism. Rather than retraining or modifying the model architecture, this mechanism leverages additional test-time compute to inject targeted prompts that improve the model's reasoning trajectory. Figure~\ref{fig:scaling_curve} illustrates how performance evolves as the number of guided reasoning steps increases, using Qwen2.5-14B-Coder evaluated on the AIME24 benchmark.

We observe a strong and consistent positive correlation between the number of interventions and the resulting accuracy. With only two guided interventions, the model's accuracy improves from 41.7\% to 45.0\%, and with four interventions, it reaches 46.7\%. These gains are achieved without any additional fine-tuning, demonstrating the utility of test-time intervention as a lightweight yet powerful means of enhancing model output. Each intervention introduces minimal overhead, yet significantly increases the model’s likelihood of correcting premature conclusions, exploring alternative strategies, and verifying results before finalizing answers.

Importantly, this improvement trend holds across multiple model architectures and reasoning domains, underscoring the generality of our intervention design. Unlike the "budget forcing" method in s1~\cite{muennighoff2025s1}, which imposes fixed computation budgets and often plateaus in effectiveness, our adaptive intervention approach dynamically responds to the model's intermediate reasoning state. This allows for more nuanced and situation-specific guidance, leading to continuous performance gains even at higher intervention counts. The results suggest that guided inference intervention is not only scalable and model-agnostic, but also opens a new axis of control for reasoning enhancement during inference—one that complements traditional training-time improvements.

\subsection{Ablation Studies}
\label{sec:ablation}

To understand the contribution of each component in our approach, we conduct comprehensive ablation studies on Qwen2.5-14B-Coder.

\begin{table}[t]
\centering
\small
\caption{\textbf{Ablation studies on ReasonBridge components.} Each row removes or modifies a specific component of our methodology to evaluate its impact.}
\resizebox{\linewidth}{!}{
\begin{tabular}{l@{\hspace{8pt}}r@{\hspace{8pt}}r@{\hspace{8pt}}r@{\hspace{20pt}}|@{\hspace{20pt}}l@{\hspace{8pt}}r@{\hspace{8pt}}r@{\hspace{8pt}}r}
\toprule
Model Variant & \makecell{AIME\\2024} & \makecell{MATH\\500} & \makecell{GPQA} & Model Variant & \makecell{AIME\\2024} & \makecell{MATH\\500} & \makecell{GPQA} \\
\midrule
Full ReasonBridge & 46.7 & 90.6 & 58.1 & w/o Strategic Adapters & 40.0 & 87.8 & 54.5 \\
\midrule
w/o Quality Filter & 36.7 & 85.4 & 51.5 & w/o Tactical Adapters & 43.3 & 88.6 & 56.1 \\
w/o Difficulty Filter & 38.3 & 86.2 & 52.0 & w/o Operational Adapters & 45.0 & 89.2 & 57.1 \\
\midrule
w/o Diversity Sampling & 33.3 & 84.8 & 50.5 & w/ Standard LoRA & 41.7 & 86.6 & 53.5 \\
\midrule
w/ Random Dataset (1K) & 33.3 & 83.4 & 49.0 & w/ Full Finetuning & 40.0 & 86.0 & 54.5 \\
w/ Full Dataset (58K) & 43.3 & 88.0 & 55.6 & w/ Simple Budget Forcing & 43.3 & 88.4 & 55.0 \\
\bottomrule
\end{tabular}}
\label{tab:ablation}
\end{table}

\subsubsection{Dataset Ablations}

The first section of Table \ref{tab:ablation} evaluates the importance of our data selection criteria. Removing the quality filter results in a 10.0 percentage point drop on AIME24, while removing the difficulty filter leads to an 8.4 point drop. The diversity sampling proves particularly crucial, with a 13.4 point drop when removed.

Comparing our carefully curated \data{} dataset with a randomly selected 1,000 examples shows the effectiveness of our selection approach, with a 13.4 point performance gap on AIME24. Interestingly, using the full 58K dataset provides only marginal benefits compared to our 1,000 selected examples, despite requiring 58× more data and substantially more training time. This confirms our hypothesis that strategic selection is more important than quantity for effective reasoning transfer, aligning with findings in s1 \cite{muennighoff2025s1} and LIMA \cite{zhou2023lima}.

\subsubsection{Architecture Ablations}

The second section of Table \ref{tab:ablation} evaluates our hierarchical adapter architecture. Strategic adapters provide the largest contribution (6.7 point drop when removed), followed by tactical adapters (3.4 points) and operational adapters (1.7 points). This aligns with our understanding that high-level reasoning strategies are particularly important for complex problem-solving.

Comparing our specialized adapters with standard LoRA and full finetuning reveals the effectiveness of our approach. Our method outperforms standard LoRA by 5.0 points on AIME24 and full finetuning by 6.7 points, while requiring far fewer trainable parameters. This demonstrates that targeted parameter modification focused on reasoning-specific capabilities is more effective than general-purpose fine-tuning techniques.

\subsubsection{Intervention Ablations}

The final row in Table \ref{tab:ablation} compares our guided inference intervention with the simple budget forcing approach used in s1 \cite{muennighoff2025s1}. Our method outperforms simple budget forcing by 3.4 points on AIME24, 2.2 points on MATH500, and 3.1 points on GPQA. This demonstrates that adaptive, content-aware interventions are more effective than uniform forcing mechanisms for scaling test-time compute.

\section{Conclusion}
\label{sec:conclusion}

We introduce ReasonBridge, a practical and efficient framework for transferring reasoning capabilities from closed-source to open-source language models via hierarchical knowledge distillation. By combining high-quality data curation, reasoning-specialized adapters, and guided inference intervention, ReasonBridge boosts reasoning performance by up to 23\% on challenging benchmarks, significantly narrowing the gap with proprietary models. Our approach requires only 1,000 curated examples and minimal parameter updates (0.3\%), offering a lightweight yet powerful solution. Through structured reasoning and test-time scaling, ReasonBridge advances the development of more capable, transparent, and widely accessible open-source AI systems.

\subsection{Limitations}
\label{sec:limitations}

While ReasonBridge is effective, it depends on closed-source models to generate initial reasoning traces, which limits accessibility. Future work could explore self-improving loops where enhanced open-source models generate and refine their own traces. Additionally, guided inference intervention increases latency due to extra reasoning steps. More efficient intervention strategies could reduce this overhead without sacrificing performance.

Our method shows strong generalization, but domain-specific performance varies. Exploring adaptive adapters or domain-aware prompting could help. We also focused on models up to 14B parameters; future work should test scalability to larger models (e.g., 70B+). Finally, combining our approach with techniques like verification-based selection~\cite{lightman2023letsverifystepstep} or tree-of-thought prompting~\cite{yao2023tree} may further enhance reasoning.
\section{Ethics Statement}

Our study aims to improve the reasoning capabilities of open-source language models by transferring knowledge from closed-source counterparts. While this transfer leverages outputs from proprietary systems, we ensure that no proprietary data or system internals are used beyond publicly accessible model outputs. The curated dataset of reasoning traces was created with careful attention to quality, diversity, and difficulty, with no personally identifiable or sensitive information involved. All benchmarks used are publicly available and designed for academic use. Our methodology supports transparency, efficiency, and reproducibility, contributing toward equitable and responsible AI development. Nevertheless, reliance on closed-source outputs raises accessibility concerns; future work should explore fully self-improving open models to mitigate this dependency.

\bibliography{main}

\appendix

\section{Example Appendix}
\label{sec:appendix}

\section{Algorithm}~\label{ag}

\begin{algorithm}[ht]
\caption{Guided Inference Intervention}
\label{alg:intervention}
\begin{algorithmic}[1]
\STATE \textbf{Input:} Problem $x$, Model $M$, Max reasoning steps $T$
\STATE \textbf{Output:} Enhanced solution $y$
\STATE $g \gets \emptyset$ \COMMENT{Initialize generation}
\STATE $t \gets 0$ \COMMENT{Initialize step counter}
\WHILE{$t < T$ \AND $\text{IsReasoningComplete}(g) = \text{False}$}
    \STATE $g \gets g \oplus M(x \oplus g)$ \COMMENT{Continue generation}
    \STATE $t \gets t + 1$
    \IF{$\text{IsTerminating}(g) = \text{True}$}
        \STATE $s \gets \text{DetectReasoningState}(g)$
        \IF{$s = \text{PARTIAL}$}
            \STATE $g \gets g \oplus$ "Wait, let me think further."
        \ELSIF{$s = \text{UNCERTAIN}$}
            \STATE $g \gets g \oplus$ "Let me try a different approach."
        \ELSIF{$s = \text{UNVERIFIED}$}
            \STATE $g \gets g \oplus$ "Let me verify this solution."
        \ELSE
            \STATE \textbf{break}
        \ENDIF
    \ENDIF
\ENDWHILE
\STATE $y \gets \text{ExtractSolution}(g)$
\STATE \textbf{return} $y$
\end{algorithmic}
\end{algorithm}

\begin{algorithm}[t]
\caption{Strategic Data Curation for \data{}}
\label{alg:datacuration}
\begin{algorithmic}[1]
\STATE \textbf{Input:} Initial pool $P$ of problem-reasoning-solution triplets
\STATE \textbf{Output:} \data{} dataset with 1,000 selected triplets
\STATE $P_{quality} \gets \emptyset$ \COMMENT{Initialize quality-filtered pool}
\FORALL{$(p, r, s) \in P$}
    \IF{$\text{HasFormattingIssues}(p, r, s) = \text{False}$ \AND $\text{IsWellStructured}(r) = \text{True}$}
        \STATE $P_{quality} \gets P_{quality} \cup \{(p, r, s)\}$
    \ENDIF
\ENDFOR
\STATE $P_{difficulty} \gets \emptyset$ \COMMENT{Initialize difficulty-filtered pool}
\FORALL{$(p, r, s) \in P_{quality}$}
    \STATE $correct_7B \gets \text{IsCorrect}(\text{Qwen2.5-7B}(p))$
    \STATE $correct_{32B} \gets \text{IsCorrect}(\text{Qwen2.5-32B}(p))$
    \IF{$correct_7B = \text{False}$ \AND $correct_{32B} = \text{False}$}
        \STATE $P_{difficulty} \gets P_{difficulty} \cup \{(p, r, s)\}$
    \ENDIF
\ENDFOR
\STATE $D \gets \emptyset$ \COMMENT{Initialize dataset}
\STATE $D_{categories} \gets \text{ClassifyIntoDomains}(P_{difficulty})$
\WHILE{$|D| < 1000$}
    \STATE $c \gets \text{SampleUniformlyFromCategories}(D_{categories})$
    \STATE $(p, r, s) \gets \text{SampleByTokenLength}(D_{categories}[c])$
    \STATE $D \gets D \cup \{(p, r, s)\}$
\ENDWHILE
\STATE \textbf{return} $D$
\end{algorithmic}
\end{algorithm}

\section{Additional experimental results}

\subsection{Implementation Details}
\label{sec:implementation2}

We implement our approach for five state-of-the-art open-source models released after 2024:

- Qwen2.5-14B-Coder \cite{Yang2024Qwen25TR}
- Qwen2.5-7B-Instruct \cite{Yang2024Qwen25TR}
- DeepSeek-7B-Coder-v1.5 \cite{guo2024deepseekcoderlargelanguagemodel}
- Yi-1.5-9B \cite{ai2025yiopenfoundationmodels}
- Mixtral-8x7B-Instruct-v0.1 \cite{jiang2023mistral}

For each model, we use a bottleneck dimension of $r = 64$ for all adapter modules, resulting in approximately 0.3\% trainable parameters compared to the full model. We distribute adapters across transformer layers based on our three-level abstraction hierarchy.

\begin{table}[ht]
\centering
\caption{\textbf{Implementation details.} Adapter configurations and training parameters for each model architecture.}
\resizebox{\linewidth}{!}{
\begin{tabular}{lrrrr}
\toprule
Model & \makecell{Adapter\\Dim} & \makecell{\# Params\\Modified} & \makecell{Learning\\Rate} & \makecell{Training\\Time} \\
\midrule
Qwen2.5-14B & 64 & 0.31\% & 5e-5 & 1.2h \\
Qwen2.5-7B & 64 & 0.33\% & 5e-5 & 0.7h \\
DeepSeek-7B & 64 & 0.32\% & 5e-5 & 0.7h \\
Yi-1.5-9B & 64 & 0.29\% & 5e-5 & 0.8h \\
Mixtral-8x7B & 64 & 0.30\% & 5e-5 & 0.9h \\
\bottomrule
\end{tabular}}
\label{tab:implementation}
\end{table}

Training was conducted using 8 NVIDIA H100 GPUs with mixed-precision (bfloat16) and gradient checkpointing for memory efficiency. We use the AdamW optimizer with a learning rate of 5e-5 and a cosine decay schedule. Training time ranged from 0.7 to 1.2 hours depending on model size, making our approach highly efficient in terms of computational resources. This aligns with findings from s1 \cite{muennighoff2025s1} that showed reasonable training times of 26 minutes on 16 H100 GPUs for their 1,000-sample approach.

\subsection{Case Study: Reasoning Pattern Analysis}
\label{sec:case_study}

To better understand how ReasonBridge enhances reasoning capabilities, we analyze the reasoning patterns of the base Qwen2.5-14B-Coder model and our enhanced version on a challenging AIME24 problem.

\begin{figure*}[t]
\centering
\includegraphics[width=\textwidth]{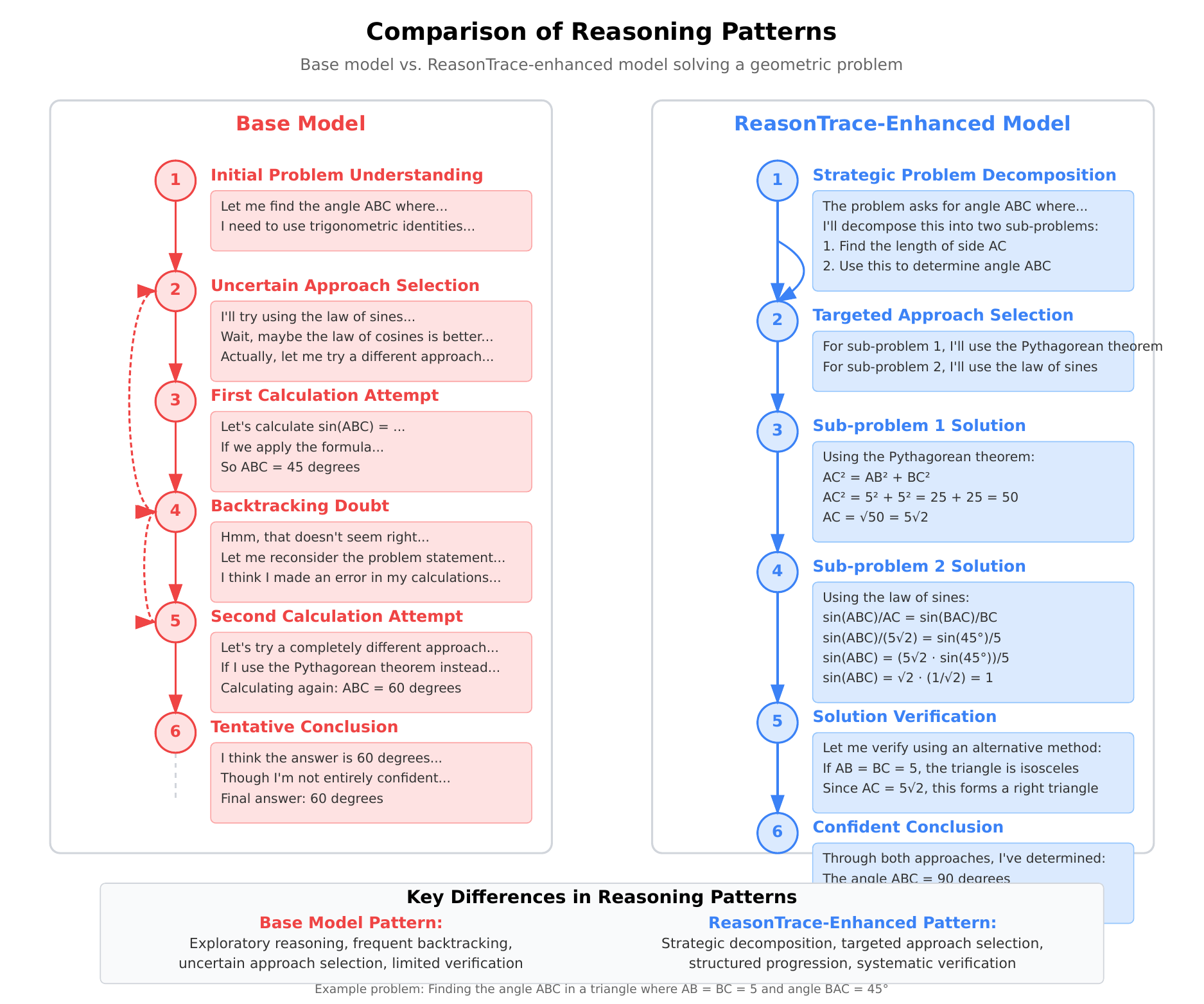}
\caption{\textbf{Comparison of reasoning patterns.} The base model exhibits frequent backtracking and uncertainty, while the ReasonBridge-enhanced version follows a more systematic approach with clearer problem decomposition and targeted solution paths.}
\label{fig:reasoning}
\end{figure*}

Figure \ref{fig:reasoning} visualizes the reasoning flows using a simplified representation where nodes represent reasoning steps and edges represent transitions between steps. Several key differences emerge:

\textbf{Problem Decomposition:} The base model attempts to solve the problem directly, leading to confusion and backtracking. In contrast, the ReasonBridge-enhanced model begins with a systematic decomposition of the problem into manageable subproblems.

\textbf{Strategic Approach Selection:} The base model tries multiple approaches in a somewhat random fashion, while the enhanced model selects appropriate techniques based on problem characteristics and follows them consistently.

\textbf{Error Recovery:} When encountering calculation errors, the base model often fails to recover, while the enhanced model employs structured verification steps that enable effective error detection and correction.

\textbf{Solution Efficiency:} The enhanced model requires fewer total reasoning steps to reach the correct solution, demonstrating more efficient problem-solving despite generating a longer overall reasoning trace.

This analysis confirms that ReasonBridge successfully transfers the structured reasoning patterns that make closed-source models effective at complex problem-solving.

\subsection{Generalization to New Tasks}
\label{sec:generalization}

\begin{table}[t]
\centering
\caption{\textbf{Generalization to diverse reasoning tasks.} Performance of ReasonBridge-enhanced Qwen2.5-14B-Coder on additional benchmarks beyond our primary evaluation targets.}
\resizebox{\linewidth}{!}{
\begin{tabular}{lrr}
\toprule
Benchmark & Base Model & + ReasonBridge \\
\midrule
MMLU (STEM) & 66.8 & 73.5 \\
GSM8K & 75.6 & 87.3 \\
BBH & 59.4 & 68.1 \\
LogiQA & 56.8 & 64.5 \\
HumanEval & 65.2 & 71.3 \\
\bottomrule
\end{tabular}}
\label{tab:generalization}
\end{table}

To assess whether ReasonBridge's benefits extend beyond our primary evaluation benchmarks, we test the enhanced Qwen2.5-14B-Coder model on five additional reasoning tasks spanning different domains:

\textbf{MMLU (STEM)} \cite{hendrycks2020measuring}: Multiple-choice questions across STEM fields.

\textbf{GSM8K} \cite{cobbe2021training}: Grade-school level math word problems.

\textbf{BBH} \cite{srivastava2023imitationgamequantifyingextrapolating}: Challenging reasoning tasks from the BIG-Bench Hard subset.

\textbf{LogiQA} \cite{liu2020logiqa}: Logical reasoning questions from admission tests.

\textbf{HumanEval} \cite{chen2021codex}: Programming problems testing code generation.

Table \ref{tab:generalization} shows consistent improvements across all tasks, with an average gain of 8.2 percentage points. The largest improvement is on GSM8K (+11.7), which involves multi-step mathematical reasoning similar to our training data. However, even on LogiQA (+7.7) and HumanEval (+6.1), which involve different reasoning domains, the gains are substantial. This demonstrates that ReasonBridge transfers generalizable reasoning capabilities rather than task-specific knowledge.

\label{sec:discussion}

\subsection{Key Insights}
\label{sec:insights}

Our experimental results yield several important insights about reasoning transfer in language models:

\textbf{Hierarchical nature of reasoning:} The effectiveness of our three-level adapter architecture suggests that reasoning operates at multiple levels of abstraction. Strategic reasoning (problem decomposition and approach selection) appears particularly important, as removing strategic adapters caused the largest performance drop in our ablation studies.

\textbf{Data efficiency in reasoning transfer:} Our results demonstrate that carefully selected data can be remarkably effective for reasoning transfer. The minimal performance gap between our 1,000-example dataset and the full 58K dataset suggests that the quality, diversity, and difficulty of examples are far more important than quantity. This aligns with findings from both s1 \cite{muennighoff2025s1} and LIMA \cite{zhou2023lima}, confirming that sample efficiency is achievable across different reasoning enhancement approaches.

\textbf{Adapter specialization benefits:} Our reasoning-specialized adapters outperformed standard LoRA and full finetuning despite modifying fewer parameters. This indicates that targeted parameter modifications focused on specific capabilities can be more effective than general-purpose fine-tuning techniques.

\textbf{Test-time compute scaling:} The consistent improvements achieved through guided inference intervention demonstrate that reasoning performance can be effectively scaled at test time through appropriate guidance. This provides a practical way to trade off compute for improved performance on challenging problems, extending upon the simpler budget forcing approach introduced in s1 \cite{muennighoff2025s1}.

\subsection{Broader Impact}
\label{sec:impact}

The ability to efficiently transfer reasoning capabilities from closed to open-source models has several potential broader impacts:

\textbf{Democratizing access:} By narrowing the performance gap between closed and open-source models, our approach helps democratize access to advanced reasoning capabilities, enabling their use in privacy-sensitive or resource-constrained environments.

\textbf{Educational applications:} Models enhanced with ReasonBridge generate more systematic and transparent reasoning traces, making them valuable tools for educational settings where explaining problem-solving approaches is as important as providing correct answers.

\textbf{Resource efficiency:} Our approach's data and parameter efficiency contributes to more sustainable AI development by reducing the computational resources required to develop high-performing models for reasoning tasks.



\newpage

\end{document}